\documentclass{article}

\usepackage{microtype}
\usepackage{graphicx}
\usepackage{subfig}
\usepackage{booktabs} % for professional tables

\usepackage{hyperref}
\usepackage{url}
\usepackage{graphicx}
\usepackage{amsmath}
\usepackage{amssymb}  % assumes amsmath package
\usepackage[accepted]{icml2018}

\icmltitlerunning{Self-Consistent Trajectory Autoencoder}

\begin{document}

\twocolumn[
\icmltitle{Self-Consistent Trajectory Autoencoder: Hierarchical Reinforcement Learning with Trajectory Embeddings}

\icmlsetsymbol{equal}{*}

\begin{icmlauthorlist}
\icmlauthor{John D. Co-Reyes}{equal,ucb}
\icmlauthor{YuXuan Liu}{equal,ucb}
\icmlauthor{Abhishek Gupta}{equal,ucb}
\icmlauthor{Benjamin Eysenbach}{goo}
\icmlauthor{Pieter Abbeel}{ucb}
\icmlauthor{Sergey Levine}{ucb}
\end{icmlauthorlist}

\icmlaffiliation{ucb}{University of California, Berkeley}
\icmlaffiliation{goo}{Google Brain}

\icmlcorrespondingauthor{John D Co-Reyes}{jcoreyes@eecs.berkeley.edu}
\icmlcorrespondingauthor{YuXuan Liu}{yuxuanliu@berkeley.edu}

\icmlkeywords{Generative Models, Hierarchical Reinforcement Learning, Unsupervised Exploration, Machine Learning}

\vskip 0.3in
]

\printAffiliationsAndNotice{\icmlEqualContribution} % otherwise use the standard text.

\begin{abstract}
In this work, we take a representation learning perspective on hierarchical reinforcement learning, where the problem of learning lower layers in a hierarchy is transformed into the problem of learning trajectory-level generative models. We show that we can learn continuous latent representations of trajectories, which are effective in solving temporally extended and multi-stage problems. Our proposed model, SeCTAR, draws inspiration from variational autoencoders, and learns latent representations of trajectories. A key component of this method is to learn both a latent-conditioned policy and a latent-conditioned model which are consistent with each other. Given the same latent, the policy generates a trajectory which should match the trajectory predicted by the model. This model provides a built-in prediction mechanism, by predicting the outcome of closed loop policy behavior. We propose a novel algorithm for performing hierarchical RL with this model, combining model-based planning in the learned latent space with an unsupervised exploration objective. We show that our model is effective at reasoning over long horizons with sparse rewards for several simulated tasks, outperforming standard reinforcement learning methods and prior methods for hierarchical reasoning, model-based planning, and exploration.
\end{abstract}

\section{Introduction}

Deep reinforcement learning (RL) algorithms can learn complex skills from raw observations~\cite{Atari, Levine16, AlphaGo}. However, domains that involve temporally extended tasks and extremely delayed or sparse rewards can pose a tremendous challenge for standard methods. A longtime goal in RL has been to develop effective hierarchy induction methods that can acquire temporally extended lower-level primitives, which can then be built upon by a higher level policy that operates at a coarser level of temporal abstraction~\cite{options,feudal,MaxQ, HAM}. A higher-level policy that is provided with temporally extended and intelligent behaviors can reason at a higher level of abstraction and solve more temporally-extended tasks. Furthermore, the same lower-level skills could be reused to accomplish multiple tasks efficiently. 

Prior work has proposed to acquire discrete sets of lower-level skills through hand-specification of objectives or bottlenecks~\cite{carlos, mlsh, options} and top-down training of hierarchically-organized policies~\cite{feudal, feudalnets}. Requiring prior knowledge and hand-specification restricts the generality of the method, while purely top-down training suffers from challenging optimization and exploration and limits the reusability of lower-level skills, providing a solution to just one task. Furthermore, the top-level meta-policy must still be trained with reinforcement learning for each task, and while this tends to be more efficient than learning from scratch if the skills are useful, it still requires considerable time and experience collection. Several works have also proposed ``bottom up'' training of lower-level skills using unsupervised objectives~\cite{optioncritic, VIC}, but such methods either also require hand-specifying some prior knowledge, or learn discrete skills that may not necessarily be sufficient to solve the higher level task. 

In this work, we propose a novel hierarchical reinforcement learning algorithm (SeCTAR) that uses a bottom up approach to learn continuous representations for trajectories, without the explicit need for hand-specification or subgoal information. Our work builds on two main ideas: first, we propose to build a continuous latent space of skills, rather than a discrete set of behaviors or options, and second, we propose to use a probabilistic latent variable model that simultaneously learns to produce skills in the world and predict their outcomes. By providing a higher-level controller with a continuous space of behaviors, it can exercise considerable control, without being restricted to a small discrete set of primitives. At the same time, since the behaviors are temporally extended, the higher-level policy still benefits from temporal abstraction. Furthermore, by training a model that both acquires a set of skills and predicts their outcomes, we can avoid needing to train a higher-level policy with reinforcement learning, and directly use these outcome predictions to perform model-based control at the higher level. This results in a hybrid model-free and model-based method, where the behaviors that actually interact with the environment are trained in model-free fashion, while the higher-level behavior is model-based. This also neatly addresses one of the major shortcomings of model-based reinforcement learning, which is the difficulty of accurately predicting low-level physical events at a fine temporal resolution. Since the predictions only need to accurately reflect the outcomes of closed-loop and temporally extended behaviors, they are substantially easier than low-level modeling of environment dynamics, while still being conducive to effective higher-level planning.

Our model is based on a trajectory-level variational autoencoder (VAE)~\cite{VAE}. The continuous latent space of behaviors is constructed by learning to embed and generate trajectories obtained via a fully unsupervised exploration objective. In addition to learning to generate the state sequences along these trajectories, the model simultaneously learns to reproduce those trajectories in the environment via a policy conditioned on the VAE latent variable. In this way, the latent-conditioned policy aims to ``imitate'' the VAE decoder. The fact that the latent-conditioned policy and the VAE decoder are representing the same behavior allows us to treat the decoder as a model of the closed loop behavior of the policy. This allows us to use the decoder to plan in the latent space by sampling latents and simulating their corresponding trajectories. We can then choose the best latents that solve the task and execute the plan with the latent-conditioned policy.

The main contribution of our work is a hierarchical reinforcement learning algorithm that acquires a continuous low-level latent space of skills, together with a predictive model that can predict the outcomes of those skills, which can be used to carry out more complex higher-level tasks. We propose a novel training procedure for this model, and show that higher-level extended tasks can be performed directly with model-based planning, without any additional reinforcement learning to learn a high level policy. Our experimental evaluation demonstrates that this approach can be used to accomplish a variety of delayed and sparse reward tasks, including interaction with objects and waypoint navigation, while outperforming reinforcement learning methods such as TRPO~\cite{TRPO}, exploration driven methods such as VIME~\cite{VIME} as well as prior work on hierarchical reinforcement learning such as FeUdal Networks~\cite{feudalnets} and option critic~\cite{optioncritic}. All our results, videos, and experimental details can be found at \url{https://sites.google.com/view/sectar}

\section{Background}
Our proposed model solves a reinforcement learning problem using components from variational inference for representation learning. The goal in reinforcement learning is to maximize the expected discounted sum of rewards:
\begin{equation}
\eta(\pi) = \mathbb{E}_{\pi}\bigg[\sum_{t=0}^H \gamma^t R(s_t, a_t)\bigg],
\end{equation}
where $\pi(a_t\mid s_t)$ is a policy that defines a distribution over actions $a_t$, $s_t$ represents the states in a Markov decision process that transition according to unknown dynamics $p(s_{t+1}\mid s_t,a_t)$, and \hbox{$R: S \times A \rightarrow R$} is a reward function. Our goal will be to solve reinforcement learning problems with long horizons and delayed rewards. Like most model-based RL methods, we assume that we have access to the reward function $R$ which we can evaluate on arbitrary states \cite{anusha, PILCO}. 

An important component of our solution is based on the framework of variational inference.
Variational inference methods use a tractable proxy distribution $q(z\mid x)$ to estimate an intractable posterior $p(z\mid x)$. Given a model with observations $x$ and latent variables $z$ we can decompose the likelihood $p(x)$ in terms of $q(z\mid x)$:
\begin{equation}
\log p(x) = D_{KL}(q(z\mid x) \; \| \;p(z\mid x)) + \mathcal{L}(x)
\end{equation}
where $\mathcal{L}(x) = \mathbb{E}_q[\log p(x\mid z)] - D_{KL}(q(z\mid x)\; \| \;p(z))$ is called the evidence lower bound  (ELBO). Since KL divergence is non-negative, we obtain the lower bound:
\begin{equation}
\begin{split}
\log p(x) &\geq \mathbb{E}_q[\log p(x\mid z)] - D_{KL}(q(z\mid x)\; \| \;p(z))
\end{split}
\end{equation}
The variational autoencoder is a particular realization of this variational inference procedure. This model can be trained by maximizing the ELBO using standard optimization methods. We refer the reader to \citet{SVI, VAE} for details.

\section{Self-Consistent Trajectory Autoencoder}

In this work, our aim is to perform long-horizon planning by learning latent representations over trajectories. Given a task with a long horizon $H$, we define trajectories in the context of SeCTAR as sequences of states $[s_0, s_1,..., s_T]$ of length $T$, where $T < H$. Each complete episode in the MDP $M$ (of length $H$) may be composed of several of these shorter trajectories. We hypothesize that building representations for these trajectories will allow us to reason more effectively over the entire horizon.

To that end, we introduce the self-consistent trajectory autoencoder (SeCTAR) to acquire latent representations of trajectories. The SeCTAR model is based on the variational autoencoder, but with two decoders: the state decoder, which decodes latent variables directly into sequences of states, and the policy decoder, which is a latent-conditioned policy capable of generating the encoded trajectory when executed in the environment. This two-headed model allows the state decoder to be a predictive model of the behavior that a policy decoder can execute in the environment.

The latent representations learned by SeCTAR can be used for planning over long episodes, by reasoning at the level of latent variables (representing extended state sequences) rather than at the level of individual states and actions. We will introduce a model-based planning algorithm based on SeCTAR in Section~\ref{sec:mpc} to perform planning in the latent space to solve long horizon tasks.

Solving tasks with sparse rewards and long horizons requires effective exploration. We show that we can improve the exploration behavior needed for hierarchical reasoning, using the SeCTAR model and an entropy based exploration objective. This results in an iterative training procedure described in Section~\ref{sec:exploration}, which we find important for performing hierarchical tasks. We first introduce the SeCTAR model, describe how it can be trained, and show its usefulness for hierarchical planning. We then describe how we can perform exploration in the loop to improve performance.  

\subsection{Graphical Model}
\label{sec:model}
\begin{figure}[t]
    \centering
    \includegraphics[width=0.4\linewidth]{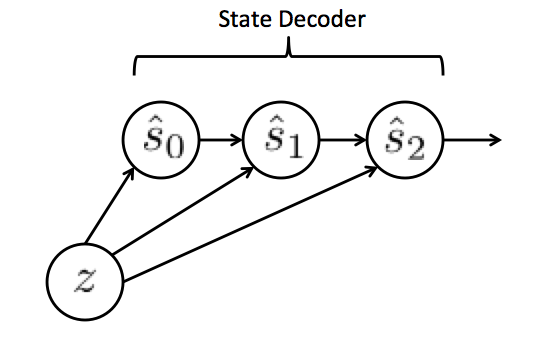} \includegraphics[width=0.25\linewidth]{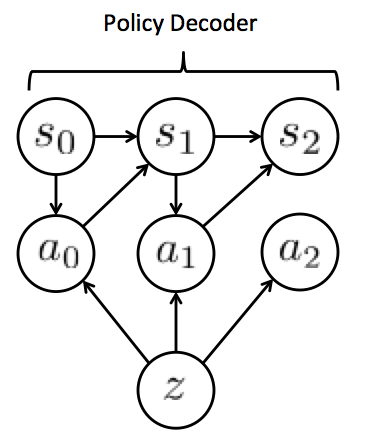}
%   \vspace{-0.3cm}
    \caption{Graphical models representing the state and policy decoders. The state decoder (shown on the left) directly generates a trajectory conditioned on the latent variable, while the policy decoder generates a trajectory by conditioning a policy which is rolled out in the environment. As is standard in model-free RL, the environment dynamics are unknown, so the policy decoder must be trained by sampling rollouts.}
  \vspace{-0.1in}
    \label{fig:graphicalmodels}
\end{figure}

We consider the problem of learning latent representations of trajectories $[s_0, s_1, \cdots, s_T]$. We begin by extending the framework of VAEs~\citep{VAE}, with trajectories $\tau$ as the observation, a trajectory-level encoder $q_{\phi}(z\mid \tau)$, and a \emph{state decoder} $p_{\theta_{SD}}(\tau\mid z)$. The graphical model representing this model is shown in Fig~\ref{fig:graphicalmodels}. We will discuss the training procedure of this model in Section~\ref{sec:training}. A trained model can generate sequences of states by sampling a latent variable $z$ and decoding using $p_{\theta_{SD}}(\tau\mid z)$.

While sequences of states are predictive of behavior, they do not allow us to act directly in the real world: the states may not be fully dynamically consistent, and we do not know the actions that would realize them. To enable our model to actually act in the world and visit states that are predicted by the state decoder $p_{\theta_{SD}}(\tau\mid z)$, we introduce a second decoder -- the policy decoder $p_{\theta_{PD}}(a\mid s,z)$. The policy decoder cannot generate the entire trajectory directly like the state decoder, but has to actually act sequentially in the environment to produce trajectories. We train this policy decoder to produce behavior in the environment consistent with the predictions made by the state decoder by minimizing the KL divergence between the distribution over state sequences under the state decoder and the policy decoder. Both the state and policy decoder are trained jointly with the recognition network $q_{\phi}(z\mid \tau)$. 

We describe the model assuming that the trajectory data $\tau$ is observed and fixed, which allows us to use maximum likelihood estimation to train the model. In Section~\ref{sec:exploration}, we will describe how we can improve trajectory distributions by alternating between model fitting and entropy based exploration, in order to generate better $\tau$ data automatically.

\subsection{Training SeCTAR with Variational Inference}
\label{sec:training}
We can train the latent variable model described in Section~\ref{sec:model} with a procedure that is similar to VAE training. Unlike a standard VAE, we must also account for the relationship between the policy decoder and state decoder. We want to maximize the likelihood of the trajectory data $p(\tau)$ under the state decoder for different $z$, while also ensuring that the state and policy decoder are consistent, minimizing the KL divergence between them. 
\begin{equation*}
\begin{aligned}
& \text{max}
& & \log p(\tau)\\
& \text{subject to}
& & \mathbb{E}_{q_{\phi}}[D_{KL}(p_{\theta_{PD}}(\tau\mid z) \; \| \; p_{\theta_{SD}}(\tau\mid z))] = 0
\end{aligned}
\end{equation*}
By applying the KL divergence as a penalty on the likelihood, we can write an unconstrained objective as 
\vspace{-0.1in}
\begin{equation}
    \max_{\theta_{SD},\theta_{PD},\phi} \log p(\tau) -\lambda \mathbb{E}_{q_{\phi}}[D_{KL}(p_{\theta_{PD}}(\tau\mid z) \; \| \; p_{\theta_{SD}}(\tau\mid z))]
\end{equation}
Introducing the evidence lower bound (ELBO) in place of the marginal likelihood $\log(p(\tau))$, we obtain
\begin{equation}\label{eq:objective}
	\begin{split}
	&\log p(\tau) -\lambda \mathbb{E}_{q_{\phi}}[D_{KL}(p_{\theta_{PD}}(\tau\mid z) \; \| \; p_{\theta_{SD}}(\tau\mid z)]\\ 
	&\hspace{0.2cm}\geq \mathbb{E}_{q_{\phi}}[\log p_{\theta_{SD}}(\tau\mid z))] - D_{KL}(q_{\phi}(z\mid \tau) \; \| \; p(z)) +\text{ }\\
	&\hspace{0.4cm}\lambda \big[\mathbb{E}_{q_{\phi}, p_{\theta_{PD}}(\tau\mid z)}[\log p_{\theta_{SD}}(\tau\mid z)] + \mathcal{H}(p_{\theta_{PD}}(\tau\mid z))\big]
  	\end{split}
\end{equation}
Intuitively, this corresponds to optimizing the ELBO while constraining the state and policy decoders to be mutually consistent. This induces the state decoder to fit the observed data and the policy decoder to match the state decoder while also maximizing the entropy of the policy's action distribution (as in maximum entropy RL \cite{schulmanmaxent}).

\begin{figure}[t]
  \centering
  \includegraphics[width=0.8\linewidth]{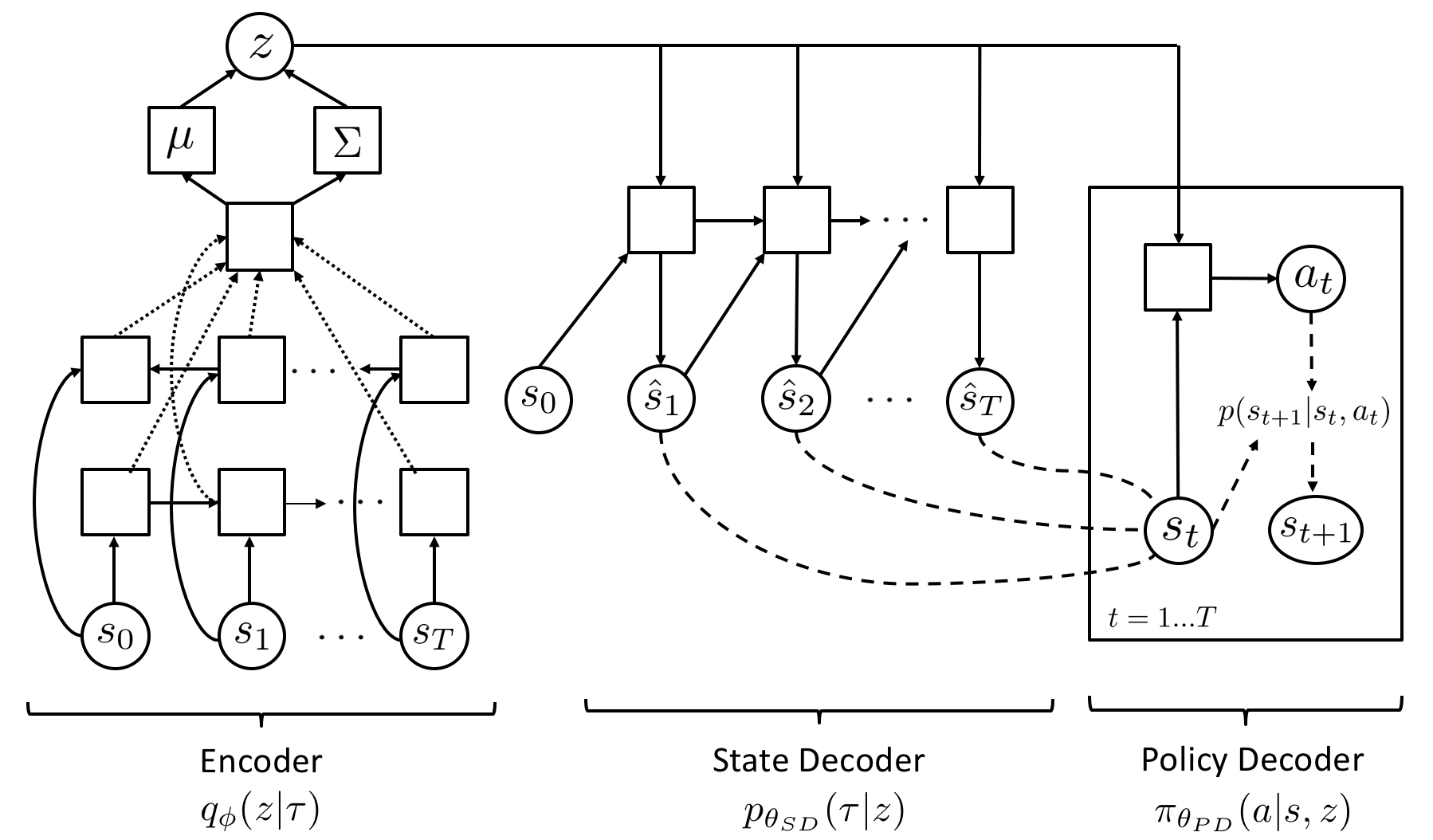}
  \caption{The SeCTAR model computation graph. A trajectory is encoded into a latent distribution, from which we sample a latent z. We then (1) directly decode z into a sequence of states using a recurrent state decoder and (2) condition a policy decoder on z to produce the same trajectory through sequential execution in the environment.}
  \label{fig:networkmodel}
\end{figure}

We parameterize our encoder $q_{\phi}(z\mid \tau)$ and state decoder $p_{\theta_{SD}}(\tau\mid z)$ with recurrent neural networks, since they operate on sequences of states, while the policy decoder is a feedforward neural network, as shown in Figure~\ref{fig:networkmodel}. Since SeCTAR will be used for generating multiple trajectories sequentially, each starting in a different state we condition the state decoder on the initial state $s_0$, allowing SeCTAR to generalize behavior across different initial states. The state decoder is completely differentiable and can be trained with backpropagation, but the policy decoder interacts with the environment's non-differentiable dynamics, so we cannot train it with backpropagation through time, instead requiring reinforcement learning.  

Optimization of the objective in Equation~\ref{eq:objective}, with respect to each of the parameters $\theta_{SD}, \theta_{PD}, \phi$ yields the different components of our model training.

\textbf{State Decoder}: Optimizing the objective with respect to $\theta_{SD}$ maximizes the terms \mbox{$\mathbb{E}_{q}[\log p_{\theta_{SD}}(\tau\mid z)] + \lambda \mathbb{E}_{q, p_{\theta_{PD}}(\tau'\mid z)}[\log p_{\theta_{SD}}(\tau'\mid z)]$.} The first term encourages the state decoder to maximize the likelihood of the observed data, while the second term encourages the state decoder to match the policy decoder. In practice, we didn't find a significant advantage in optimizing the second term with respect to $\theta_{SD}$ so it is omitted from our implementation. Since $\theta_{SD}$ is differentiable this objective can be directly optimized using backpropagation. 

\textbf{Policy Decoder}: Optimizing with respect to $\theta_{PD}$ maximizes the terms \mbox{$\lambda \big[\mathbb{E}_{q, p_{\theta_{PD}}(\tau'\mid z)}[\log p_{\theta_{SD}}(\tau'\mid z)] + \mathcal{H}(p_{\theta_{PD}}(\tau\mid z))\big]$}. The first term encourages samples drawn from the policy decoder to maximize the likelihood under the state decoder, while the second term is an entropy regularization. Since $p_{\theta_{PD}}(\tau\mid z)$ is non differentiable, we use reinforcement learning to optimize this objective with reward computed by trajectory likelihood under the state decoder, regularized with an entropy objective. In practice, trajectory data from the environment actually consists of sequences of both states \emph{and} actions. We find that pretraining the policy decoder with behavior cloning to match the actions in the trajectory provides a good initialization for subsequent finetuning with RL. 

To optimize this model, we sample a batch of trajectories from the current set of training trajectories and alternate between training the state decoder with backpropagation with the standard VAE loss and training the policy decoder by initializing with behavior cloning and doing RL finetuning with the reward function described above using PPO~\cite{ppo}, backpropagating gradients into the encoder in both cases. 

\subsection{Hierarchical Control with SeCTAR}
\label{sec:mpc}

After training the SeCTAR model as described above, we can apply it to perform hierarchical control. Since SeCTAR provides us with a latent representation of trajectories, we can design a meta-controller that reasons sequentially in the space of these latent variables at a coarser time scale than the individual time steps in the environment. Decision making in the latent space serves two purposes. First, it allows for more coherent exploration than randomized action selection. Second, it shortens the effective horizon of the problem to be solved in latent space.

To perform temporally extended planning, we can use a meta-controller that sequentially chooses latent space values $z$. Each latent $z$ is used to condition the policy decoder $\pi_{\theta_{PD}}(a\mid s,z)$, which is executed in the environment for $T$ steps, after which the meta-controller picks another latent. Although there are several choices for designing or learning such a meta-controller, we consider an approach using model-based planning with model predictive control (MPC), which takes advantage of the state decoder. Model predictive control is an effective control method which performs control by finite horizon model based planning, with iterative replanning at every time step. We refer readers to \cite{MPCTheory} for a comprehensive overview. 

An important property of the SeCTAR model is that the differentiable state decoder and the non-differentiable policy decoder are trained to be consistent with each other (Equation~\ref{eq:objective}). The state decoder represents a model of how the policy decoder will actually behave in the environment for a particular latent. This is similar to a dynamics model, but built at the trajectory level rather than the the transition level (i.e., operating on ($s_t$, $a_t$, $s_{t+1}$). In this work, we use this interpretation of the state decoder as a model to build a model predictive controller in latent space. Note that the state decoder only needs to make predictions about the outcomes of the corresponding closed-loop policy, which is significantly easier than forward dynamics prediction for arbitrary actions. We use the latent space as the action space for MPC, and perform simple shooting-based planning via random sampling and replanning to generate a sequence of latent variables that maximize a given reward function. 

Specifically, given an episode of length $H$ and SeCTAR trained with trajectories of length $T$, we solve the following planning problem in the latent space over a horizon of $H/T$ (the effective horizon in latent space)
\begin{equation*}
\begin{aligned}
\underset{z_0, z_1 ..., z_{H/T}}{\text{max}}
\hspace{0.2cm}& \sum_{t=0}^{H/T} \gamma ^{tT}R(\tau_t)\\
\text{subject to }\hspace{0.2cm}
& \tau_t \sim p_{\theta_{SD}}(\tau\mid z_t, s_0^t) \\ & s_0^0 = s_0, s_0^{t+1} = s_T^{t} \\
\end{aligned}
\end{equation*}
Here, $\tau_t$ is a trajectory sampled from the state decoder $p_{\theta_{SD}}(\tau\mid z_t, s_0^t)$ conditioned on the current state, and $s_0^t$ represents the start of the trajectory segment, which is the last state in the previous segment. $R(\tau_t) = \sum_{i=0}^{T-1}\gamma^iR(s^t_i, a^t_i)$ is the discounted sum of rewards of trajectory $\tau_t$.
To perform this optimization, we use a simple shooting based method~\cite{anusha} for model-based planning in latent space, described in Algorithm~\ref{alg:mpc}.

\begin{algorithm}[!t]
\begin{algorithmic}[1]
 \STATE{\textbf{Given}: trained SeCTAR model, Reward function R}
\FOR{timestep $t \in \{1,\dots,H/T\}$}
    \STATE{Sample $K$ sequences of latents from the prior where each sequence has $H_{MPC}$ number of latents}
    \STATE{Use the state decoder to predict environment states of length $T\times H_{MPC}$ for each latent sequence.}
     \STATE{Evaluate the reward per sequence, and choose the best sequence of latents.}
    \STATE{Execute the policy decoder $\pi_{\theta_{PD}}(a\mid s,z)$ conditioned on the first latent $z$ from the chosen sequence, for $T$ steps starting at $s_0^t$.}
\ENDFOR
\end{algorithmic}
\caption{Model predictive control in latent space}
\label{alg:mpc}
\end{algorithm}

\subsection{Exploration for SeCTAR}
\label{sec:exploration}

The model proposed in Section~\ref{sec:model}, provides an effective way to learn representations for trajectories and generate behavior via a state and policy decoder. However, if we assume that the trajectory data is observed and fixed as we have thus far, the trajectories that our model can generate are restricted by the distribution of observed data. This is particularly problematic in the setting of RL problems over long horizons, where there is a need to explore the environment significantly. The distribution of trajectories that SeCTAR is trained on cannot simply be fixed but needs to be updated periodically to explore more of the state space. 

In order to collect data to train the SeCTAR, we introduce a policy $\pi_e$ that we refer to as the explorer policy. The goal of the explorer policy is to collect data which is as useful as possible for training the SeCTAR model and performing hierarchical planning with it. The explorer policy should gather data by (1) exploring in regions which are relevant to the hierarchical task being solved, and (2) exploring diverse behavior within these regions.

We explore in the neighborhood of task relevant states by initializing the explorer policy near the distribution of states visited by the MPC controller described in Section~\ref{sec:mpc}. We can achieve this by running the hierarchical controller with a randomly truncated horizon, and letting the explorer policy take over execution. For environments that allow resets to a given state, we can also start the explorer policy directly from a random sample of states visited by the MPC controller.

\begin{algorithm}[t]
\begin{algorithmic}[1]
 \STATE{Initialize replay buffer and SeCTAR with data from randomly initialized $\pi_e$}
\FOR{iteration $j \in \{1,\dots,J\}$}
    \STATE{Execute model predictive control in latent space as in Algorithm~\ref{alg:mpc}}
    \STATE{Run the explorer $\pi_e$ for $T\times H_e$ starting from a random sample of states visited by MPC}
    \STATE{Update $\pi_e$ using PPO with reward as negative ELBO (\ref{eq:negelbo}) estimated on each of the $H_e$ trajectories}
    \STATE{Train SeCTAR as described in Section~\ref{sec:training} using data collected by $\pi_e$ in this iteration, mixed with some data from prior iterations in the replay buffer}
\ENDFOR
\end{algorithmic}
\caption{Overall algorithm overview}
\label{alg:fullalgorithm}
\end{algorithm}
For $\pi_e$ to explore diverse behavior, we propose maximizing the entropy of the marginal trajectory distribution $p(\tau)$ induced under $\pi_e$. Previous work on maximum entropy RL~\cite{softQ, a3c, schulmanmaxent} typically maximize the conditional entropy $\mathcal{H}(\pi(a \mid s))$ of the policy distribution $\pi(a \mid s))$. In this work we suggest maximizing the marginal entropy over distributions of entire trajectories, which is different from maximizing entropy over the policy distribution. The objective can be written as: 
\begin{equation}
		\max_{\theta} \mathcal{H}(p_{\theta}(\tau)) =-\mathbb{E}_{p_{\theta}(\tau)}[\log p_{\theta}(\tau)]
\end{equation}
Optimizing this objective reduces (on applying product rule, and removing a constant baseline) to policy gradient, with $-\log p_{\theta}(\tau)$ as the reward function per trajectory. The log likelihood $\log p_{\theta}(\tau)$ is typically intractable to estimate. However, SeCTAR provides us an effective way to estimate $\log p_{\theta}(\tau)$ by using a lower bound. SeCTAR optimizes the evidence lower bound (ELBO) to maximize likelihood of trajectories, which suggests a simple approximation for $\log p_{\theta}(\tau)$ via the negated ELBO
\begin{equation}\label{eq:negelbo}
 -\mathbb{E}_q[\log p_{\theta_{SD}}(\tau\mid z)] + D_{KL}(q(z\mid\tau) \; \| \; p(z)),
\end{equation}
as an approximation of $-\log(p_{\theta}(\tau))$. We can then perform policy gradient for exploration with this reward function. 

We combine the previously discussed model-predictive control and entropy maximization methods into an iterative procedure which interleaves exploration with model fitting and hierarchical planning, as summarized in Algorithm~\ref{alg:fullalgorithm}.

\section{Related Work}
Hierarchical reinforcement learning is a well studied area in reinforcement learning ~\cite{options, feudal, schmidhuber, HAM, MaxQ}. One method is the options framework which involves learning temporally extended subpolicies. However, the number of options is usually both finite and fixed beforehand which may not be optimal for more complex domains such as continuous control tasks. Another challenge is acquiring skills autonomously which previous work bypasses by hand engineering subgoals ~\cite{options} or using pseudo-rewards ~\cite{MaxQ}. Some end-to-end gradient-based methods to learn options have recently been proposed as well ~\cite{optioncritic, DDO}. Our work on the other hand, learns a continuous set of skills without supervision by learning representations over trajectories, and optimizing the entropy over trajectory distributions to encourage a diverse and useful set of primitives.

In most environments, good exploration is a prerequisite for hierarchy. A number of prior works have been proposed to guide exploration based on criteria such as intrinsic motivation ~\cite{schmidhuber, bradly_acvp}, state-visitation counts ~\cite{MBIE-EB, pseudocounts}, and optimism in the face of uncertainty~\cite{R-max}. In this work, we suggest a simple unsupervised exploration method which aims to maximize entropy of the marginal of trajectory distributions. This can be thought of as a means of density based exploration, related to ~\cite{pseudocounts, EX2} but operating at a trajectory level. 

Several recent and concurrent works have proposed methods which are related to ours but have clear distinctions. \citet{carlos, heess, hausman} learn stochastic neural networks to modulate low level behavior which is trained on a ``proxy'' reward function. However, our method does not assume that such a proxy reward function is provided, as it is often restrictive and difficult to obtain in practice. \citet{trajsegment} uses trajectory segment models for planning but has no mechanism for exploration and does not consider hierarchical tasks. Other works present information-theoretic representation learning frameworks that are also based on latent variable models and variational inference, but have significant differences in their methods and assumptions~\cite{VIC, variationalempowerment}. \citet{VIC} aims to learn a maximally discriminative set of options by maximizing the mutual information between the final state reached by each of the options and the latent representation. Whereas this prior method is applied only on relatively simple gridworlds with discrete options, we learn a continuous space of primitives, together with a state decoder that can be used for model-based higher-level control.

\section{Experiments}
In our experimental evaluation, we aim to address the following questions: (1) Can we learn good exploratory behavior in the absence of task reward, using SeCTAR with our proposed exploration method? (2) Can we use the learned latent space with planning and exploration in the loop to solve hierarchical and sparse reward tasks? (3) Does the state decoder model make meaningful predictions about the outcomes of the high-level actions?
We evaluate our method on four different domains: 2D navigation, object manipulation, wheeled locomotion, swimmer navigation which are shown in Figure \ref{fig:envs}. Details of the experimental evaluation can be found in the appendix.

\subsection{Tasks}

\begin{figure}[t]
    \centering
    \includegraphics[width=0.273\linewidth]{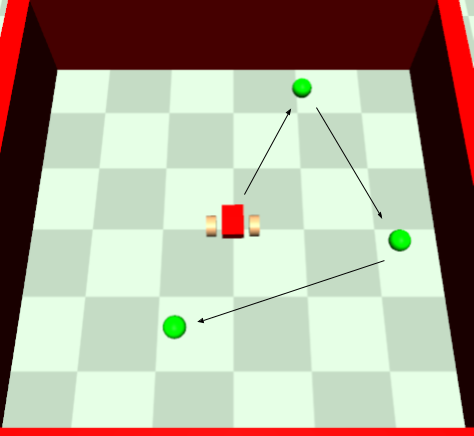}
    \includegraphics[width=0.275\linewidth]{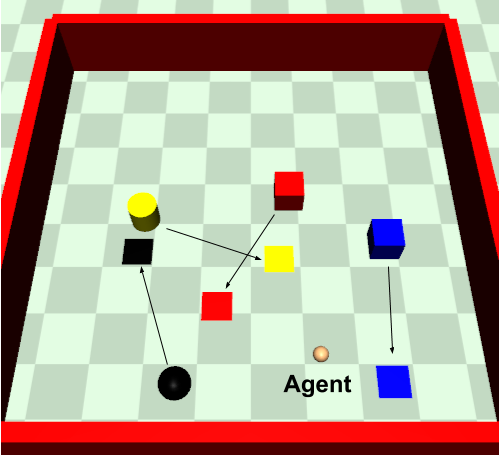}
    \includegraphics[width=0.25\linewidth, height=0.25\linewidth]{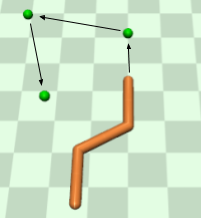}
    \vspace{-0.1in}
    \caption{From left to right (1) the wheeled locomotion environment with the waypoints depicted in green (2) the object manipulation environment with different objects (blocks and cylinders) and their correspondingly colored goals (squares) (3) the swimmer navigation task with the first 3 waypoints depicted in green.}
    \vspace{-0.2in}
    \label{fig:envs}
\end{figure}

\paragraph{2-D Navigation}
In the 2-D navigation task, the agent can move a fixed distance in each of the four cardinal directions. States are continuous and are observed as the 2D location of the agent. The objective is to navigate a specific sequence of $M$ goal waypoints which lie within a bounding box. The agent is given a reward of 1 for successfully visiting every third goal in the sequence. This evaluates our model's ability to reason over long-horizons with sparse rewards.
\vspace{-0.1in}
\paragraph{Wheeled Locomotion}
The wheeled environment consists of a two-wheeled cart that is controlled by the angular velocity of its wheels. The cart uses a differential drive system to turn and move in the plane. States include the position, velocity, rotation, and angular velocity of the cart. In this task, the cart must move to a series of goals within a bounding box and receives a reward of 1 after reaching every third goal in the sequence. This experiment tests our method's effectiveness in reasoning over a continuous action space with more complicated physics.
\vspace{-0.1in}
\paragraph{Object Manipulation}
The object manipulation environment consists of four blocks that the agent can move. The agent, which moves in 2D, can pick up nearby blocks, drop blocks, and navigate in the four cardinal directions, carrying any block it has picked up. The agent must move each block to its corresponding goal in the correct sequence and is given a reward of 1 for each correctly placed block. We designed this task to evaluate our method's ability to explore and learn useful interaction skills with objects in the environment. The sparse, sequential and discontinuous nature of this task makes it challenging.
\vspace{-0.1in}
\paragraph{Swimmer Navigation}
This task involves navigating through a number of waypoints in the correct order using a 3-link robotic swimmer. The agent is given a reward of 1 for successfully visiting every third goal. This task requires acquiring both a low-level swimming gait and a higher-level navigation strategy to visit the waypoints, and presents a more substantial exploration challenge.

\subsection{Unsupervised Exploration with SeCTAR}
To evaluate the effectiveness of the exploration method described in Section~\ref{sec:exploration}, we consider an unsupervised setting where we interact with environments in the absence of a task reward. We evaluate a simplified version of Algorithm~\ref{alg:fullalgorithm} which alternates between (1) exploration with the explorer policy $\pi_e$, (2) model fitting with SeCTAR, (3) updating $\pi_e$ via the ELBO as described in Section~\ref{sec:exploration}. This is a version of Algorithm~\ref{alg:fullalgorithm}, with no MPC and $\pi_e$ initialized at a fixed initial state.

\begin{figure}[ht]
    \centering
    \includegraphics[width=\linewidth]{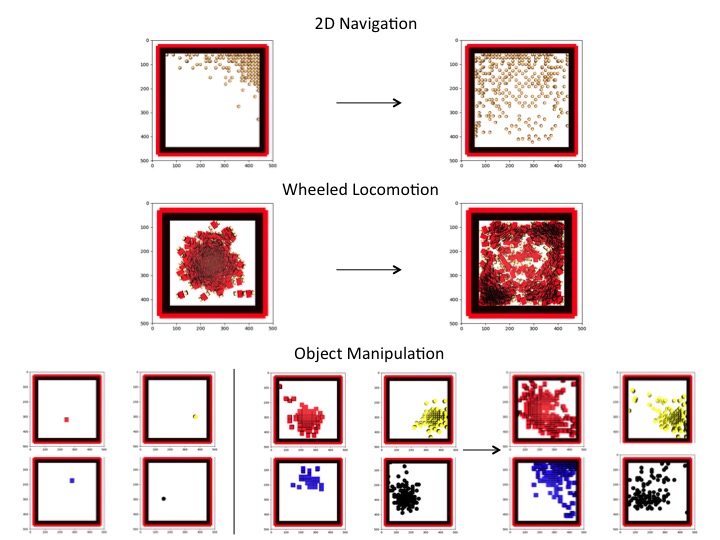}
    \caption{We show how our method improves exploration on three environments. On the left, we show the final agent locations for 2D navigation and wheeled location and show final block positions of 4 blocks for object manipulation from a randomly initialized policy. On the right we show the corresponding final locations from our explorer policy trained with the unsupervised exploration objective in Section~\ref{sec:exploration}. The bottom left plot shows the initial block positions. In all environments we see the agent learns to explore a more evenly distributed region of the state space.
    }
    \vspace{-0.1in}
    \label{fig:exploration}
\end{figure}

Our goal is to determine if alternating between exploration and SeCTAR model fitting~\ref{sec:training} provides us with effective exploration behavior, which is a prerequisite for hierarchical reinforcement learning. To evaluate this, we compare the distribution of final states visited by a randomly initialized policy and the explorer policy after unsupervised training. We found that the distribution of states of the explorer policy $\pi_e$ covered a significantly larger portion of the state space, indicating good exploratory behavior as seen in Figure~\ref{fig:exploration}. For the object manipulation task, the manipulator learns to pick up objects and move them around maximally while in the locomotion and 2D navigation environments, the agent learns to explore different portions of its state space. 

\subsection{Hierarchical Control}
For the next experiment, we compare our full Algorithm \ref{alg:fullalgorithm} against several baselines methods for exploration, hierarchy, and model-based control. To provide a fair comparison, we initialize all methods from scratch, assuming no prior training in the environment. For each environment, we randomly generated 5 sets of goal configurations and compare the average reward over all goal configurations.

We compare against model-free RL methods, TRPO \citep{TRPO} and A3C \citep{a3c}, an exploration method based on intrinsic motivation - VIME \cite{VIME}, a model-based method from \citet{anusha}, and two hierarchical methods, FeUdal Networks \cite{feudalnets} and option-critic \cite{optioncritic}. For the model-based baseline, we perform the same number of random rollouts as our method, with the same planning horizon. However, due to the computational demand of planning at every time-step, we replan at the same rate as our method. We augment the state of the environment with a one-hot encoding of the goal index to enable memoryless policies to operate effectively. We did not evaluate FeUdal and A3C on the wheeled locomotion and the swimmer navigation task, as our implementations of these methods only accommodated discrete actions.
\begin{figure}[t]
    \centering
      \includegraphics[width=0.8\linewidth]{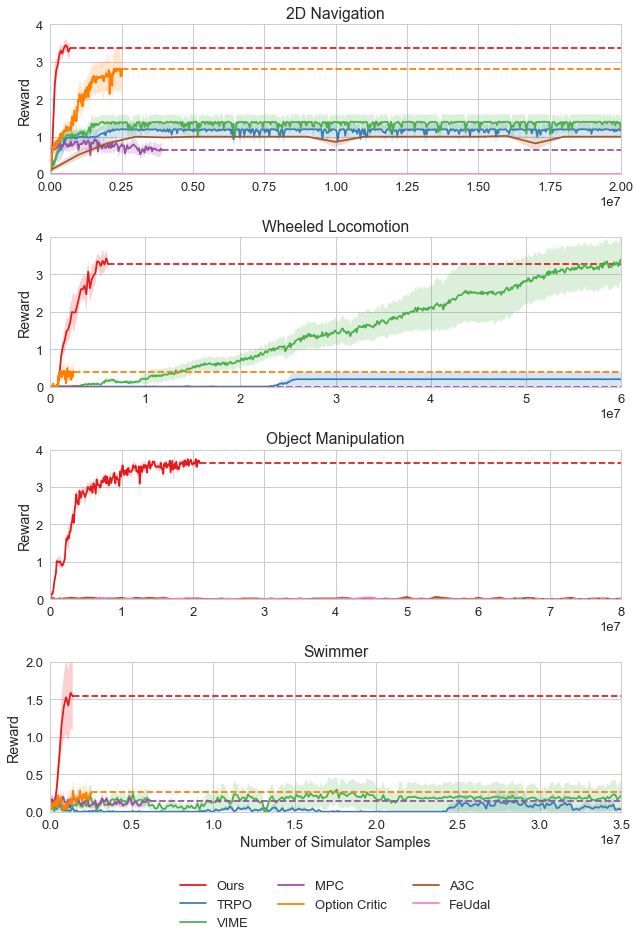}
    \caption{Comparison of our method with prior methods on the four tasks. Dashed lines indicate truncated execution. We find that on all tasks, our method is able to achieve higher reward much quicker than model-based, model-free and hierarchical baselines. For object manipulation and swimmer, prior methods fail to do anything meaningful.}
    \label{comparison}
    \vspace{-0.5cm}
\end{figure}

We found that our method can significantly outperform prior methods in terms of task performance and sample complexity as shown in Figure \ref{comparison}. These tasks require sequential long horizon reasoning and handling of delayed and sparse rewards. The block manipulation task is particularly challenging for all methods, since it requires the exploration process to pick up blocks and move them around, and only receives a reward when the blocks are placed in the correct locations sequentially. We found that our method is able to significantly outperform the model-based baseline, indicating the usefulness of building trajectory-level models, rather than predictive models at the state-action level. This is likely because model-based predictions at the trajectory level are less susceptible to compounding errors, and are only required to solve the simpler task of predicting the outcomes of specific closed-loop skills, rather than arbitrary actions.
\begin{figure}[!t]
    \centering
      \includegraphics[width=.9\linewidth]{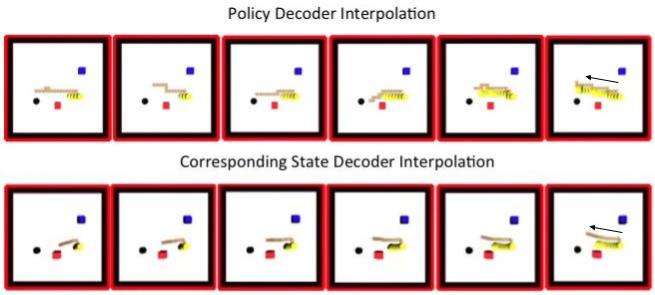}
        \caption{Interpolation between two latent codes on the object manipulation environment. We interpolate between two latent codes and visualize the corresponding trajectories from the policy decoder and the state decoder where each plot is a single trajectory. The agent position is in brown and the object positions are in blue, yellow, black and red.  From left to right, there is a smooth interpolation between moving the yellow object a little to the left and moving it much further left.}
\vspace{-0.25in}
\label{interpolation}
\end{figure}
We also found that our method performed better than TRPO, A3C, VIME, option-critic, and FeUdal Networks on all tasks.
The ability of SeCTAR to learn better on tasks which require challenging exploration and long-horizon reasoning can likely be attributed to being able to perform long-horizon planning using good trajectory representations. The model-based planner at the high level reduces sample complexity significantly, while temporally extended trajectory representations allow us to reason more effectively over longer horizons. While we find that, in the wheeled robot environment, using VIME eventually matches the performance of our method, we are significantly more sample efficient with model-based high-level planning. On the harder object manipulation and swimmer tasks, only our method achieves good performance.

\subsection{Model Analysis}
We visualize interpolations in latent space to see how well the model generalizes to unseen trajectories in Figure \ref{interpolation}. We choose a latent in the dataset and interpolate to a random point in the latent space. For each interpolated latent we visualize the predicted trajectory from the state decoder and the rolled out trajectory from the policy decoder by plotting the position of the agent. The trajectories are mostly consistent with each other, which demonstrates the potential of SeCTAR to generalize its consistency to new behavior and provide a structured and interpretable latent space.

\vspace{-0.1in}
\section{Conclusion}
We proposed a method for hierarchical reinforcement learning that combines representation learning of trajectories with model-based planning in a continuous latent space of behaviors. We describe how to train such a model and use it for long horizon planning, as well as for exploration. Experimental evaluations show that our method outperforms several prior methods and flat reinforcement learning methods in tasks that require reasoning over long horizons, handling sparse rewards, and performing multi-step compound skills.

\section{Acknowledgements}
We would like to thank Roberto Calandra, Gregory Kahn, Justin Fu for helpful comments and discussions. This work was supported by the AWS Program for Research and Education, equipment donations from NVIDIA, Berkeley Deep Drive, ONR PECASE N000141612723, and an ONR Young Investigator Program award.

\clearpage
\bibliography{references}
\bibliographystyle{icml2018}

\clearpage
\appendix

\twocolumn[
\icmltitle{Supplementary Materials}
\vskip 0.3in
]

\section{Experimental Details}
For all experiments, we parameterize $\pi_{\theta_{PD}}$ and $\pi_e$ as a three-layer fully connected neural networks with 400, 300, 200 hidden units and ReLU activations. The policies output either categorical or Gaussian distributions. The encoder is a two-layer bidirectional-LSTM with 300 hidden units, and we mean-pool over LSTM outputs over time before applying a linear transform to produce parameters of a Gaussian distribution. We use an 8-dimensional diagonal Gaussian distribution for $z$. The state decoder is a single-layer LSTM with 256 hidden units that conditions on the initial state and latent $z$, to output a Gaussian distribution over trajectories. We use trajectories of length $T=19$, and plan over $K=2048$ random latent sequences. We use horizons $H = 380$, $H_{MPC}=5$, $H_e = 5$ for the 2D navigation task, $H=950$, $H_{MPC}=20$, $H_e = 10$ for the wheeled locomotion task, and $H=950$, $H_{MPC}=10$, $H_e = 10$ for the object manipulation task. These values were chosen empirically with a hyperparameer sweep. 

\section{Baseline Details}
\paragraph{TRPO / VIME}
We used the rllab TRPO implementation, OpenAI VIME implementation with a batch size of 100 * task horizon and step size of 0.01.
\paragraph{MPC}
We use a learning rate of 0.001 and batch size of 512. The MPC policy simulates 2048 paths each time it is asked for an action. We verified correctness on half-cheetah.
\paragraph{Option Critic}
We use a version of Option Critic that uses PPO instead of DQN. We swept over number of options, reward multiplier, and entropy bonuses. We verified correctness on cartpole, hopper, and cheetah.
\paragraph{Feudal / A3C}
The Feudal and A3C implementations are based on chainerRL. We swept over the parameters $\beta$, $t_{max}$, and gradient clipping. 
\bibliographystyle{icml2018}

\end{document}

\twocolumn[
\icmltitle{Supplemental Materials}



\icmlsetsymbol{equal}{*}

\begin{icmlauthorlist}
\icmlauthor{John D. Co-Reyes}{equal,ucb}
\icmlauthor{YuXuan Liu}{equal,ucb}
\icmlauthor{Abhishek Gupta}{equal,ucb}
\icmlauthor{Benjamin Eysenbach}{goo}
\icmlauthor{Pieter Abbeel}{ucb}
\icmlauthor{Sergey Levine}{ucb}
\end{icmlauthorlist}

\icmlaffiliation{ucb}{University of California, Berkeley}
\icmlaffiliation{goo}{Google Brain}

\icmlcorrespondingauthor{John D Co-Reyes}{jcoreyes@berkeley.edu}
\icmlcorrespondingauthor{YuXuan Liu}{yuxuanliu@berkeley.edu}

\icmlkeywords{Generative Models, Hierarchical Reinforcement Learning, Unsupervised Exploration, Machine Learning}
\vskip 0.3in
]

\printAffiliationsAndNotice{\icmlEqualContribution} % otherwise use the standard text.

\section{Experimental Details}
For all experiments, we parameterize $\pi_{\theta_{PD}}$ and $\pi_e$ as a three-layer fully connected neural networks with 400, 300, 200 hidden units and ReLU activations. The policies output either categorical or Gaussian distributions. The encoder is a two-layer bidirectional-LSTM with 300 hidden units, and we mean-pool over LSTM outputs over time before applying a linear transform to produce parameters of a Gaussian distribution. We use an 8-dimensional diagonal Gaussian distribution for $z$. The state decoder is a single-layer LSTM with 256 hidden units that conditions on the initial state and latent $z$, to output a Gaussian distribution over trajectories. We use trajectories of length $T=19$, and plan over $K=2048$ random latent sequences. We use horizons $H = 380$, $H_{MPC}=5$, $H_e = 5$ for the 2D navigation task, $H=950$, $H_{MPC}=20$, $H_e = 10$ for the wheeled locomotion task, and $H=950$, $H_{MPC}=10$, $H_e = 10$ for the object manipulation task. These values were chosen empirically with a hyperparameer sweep. 

\section{Baseline Details}
\paragraph{TRPO / VIME}
We used the rllab TRPO implementation, OpenAI VIME implementation with a batch size of 100 * task horizon and step size of 0.01.
\paragraph{MPC}
We use a learning rate of 0.001 and batch size of 512. The MPC policy simulates 2048 paths each time it is asked for an action. We verified correctness on half-cheetah.
\paragraph{Option Critic}
We use a version of Option Critic that uses PPO instead of DQN. We swept over number of options, reward multiplier, and entropy bonuses. We verified correctness on cartpole, hopper, and cheetah.
\paragraph{Feudal / A3C}
The Feudal and A3C implementations are based on chainerRL. We swept over the parameters $\beta$, $t_{max}$, and gradient clipping. 
\bibliographystyle{icml2018}